\documentclass{article}
\usepackage{ijcai25}

\usepackage{times}
\usepackage{soul}
\usepackage{url}
\usepackage[hidelinks]{hyperref}
\usepackage[utf8]{inputenc}
\usepackage[small]{caption}
\usepackage{graphicx}
\usepackage{amsmath}
\usepackage{amsthm}
\usepackage{booktabs}
\usepackage{algorithm}
\usepackage{algorithmic}
\usepackage[switch]{lineno}

\usepackage{amsfonts}
\usepackage{dsfont}
\usepackage{tabularx}
\usepackage{colortbl}
\usepackage{natbib}
\usepackage{subcaption}

\title{Reward Models in Deep Reinforcement Learning: A Survey}

\author{
Rui Yu \and
Shenghua Wan \and
Yucen Wang \and
Chen-Xiao Gao \and\\
Le Gan \and
Zongzhang Zhang \and
De-Chuan Zhan \\
\affiliations
National Key Laboratory for Novel Software Technology, Nanjing University, China\\
School of Artificial Intelligence, Nanjing University, China
\emails
\{yur,wansh,wangyc,gaocx\}@lamda.nju.edu.cn,
\{ganle,zzzhang,zhandc\}@nju.edu.cn
}

\begin{document}

\maketitle

\begin{abstract}
In reinforcement learning (RL), agents continually interact with the environment and use the feedback to refine their behavior. To guide policy optimization, reward models are introduced as proxies of the desired objectives, such that when the agent maximizes the accumulated reward, it also fulfills the task designer's intentions. Recently, significant attention from both academic and industrial researchers has focused on developing reward models that not only align closely with the true objectives but also facilitate policy optimization. In this survey, we provide a comprehensive review of reward modeling techniques within the deep RL literature. We begin by outlining the background and preliminaries in reward modeling. Next, we present an overview of recent reward modeling approaches, categorizing them based on the source, the mechanism, and the learning paradigm. Building on this understanding, we discuss various applications of these reward modeling techniques and review methods for evaluating reward models. Finally, we conclude by highlighting promising research directions in reward modeling. Altogether, this survey includes both established and emerging methods, filling the vacancy of a systematic review of reward models in current literature. 
\end{abstract}

\section{Introduction}

In recent years, deep reinforcement learning (DRL), a machine learning paradigm that combines RL with deep learning, has demonstrated its immense potential in applications across various domains. For example, AlphaGo \citep{silver2016mastering} showcased RL’s capability of complex decision-making in game scenarios; InstructGPT \citep{ouyang2022training} marked the irreplaceable role of RL in aligning language models with human intents; agents trained via large-scale RL, such as OpenAI-o1 and DeepSeek-R1 \citep{guo2025deepseek}, demonstrated impressive reasoning intelligence that is comparable or even exceeds human capability. Unlike supervised learning (SL) where the agent is required to imitate and replicate the behavior recorded in the dataset, RL sets itself apart by enabling the agent to explore, adapt, and optimize its behavior based on the outcome of its actions, thereby achieving unprecedented levels of autonomy and capability.

A key component of reinforcement learning is the \textbf{reward}, which essentially defines the goal of interest in the task and guides the agents to optimize their behavior toward that intent~\citep{sutton1998reinforcement}. Just as dopamine motivates and reinforces adaptive actions in biological systems, rewards in RL encourage exploration of the environment and guide intelligent agents towards desired behaviors \citep{glimcher2011understanding}. However, while rewards are typically predefined in research environments \citep{towers2024gymnasium}, they are often absent or difficult to specify in many real-world applications. In light of this, a significant portion of modern RL research focuses on how to extract effective rewards from various types of feedback, after which standard RL algorithms can be applied to optimize the policies of agents. 

Despite the crucial role of reward modeling in RL, existing surveys \citep{arora2021survey, kaufmann2023survey} are often embedded within specific subdomains such as inverse reinforcement learning (IRL) and reinforcement learning from human feedback (RLHF), with a limited focus on reward modeling as a standalone topic. To bridge this gap, we provide a systematic review of reward models, covering their foundations, key methodologies, and applications across diverse RL settings. We introduce a new categorization framework that addresses three fundamental questions: (1) \textit{The source}: Where does the reward come from? (2) \textit{The mechanism}: What drives the agent’s learning? (3) \textit{The learning paradigm}: How to learn the reward model from various types of feedback? Furthermore, we highlight recent advancements in reward models based on foundation models, such as large language models (LLMs) and vision-language models (VLMs), which have received relatively little attention in previous surveys. The framework of reward modeling we establish in this survey is illustrated in Figure \ref{fig:pipeline}. Specifically, this survey is organized as follows:

\begin{figure*}
\centering\includegraphics[width=1.0\linewidth]{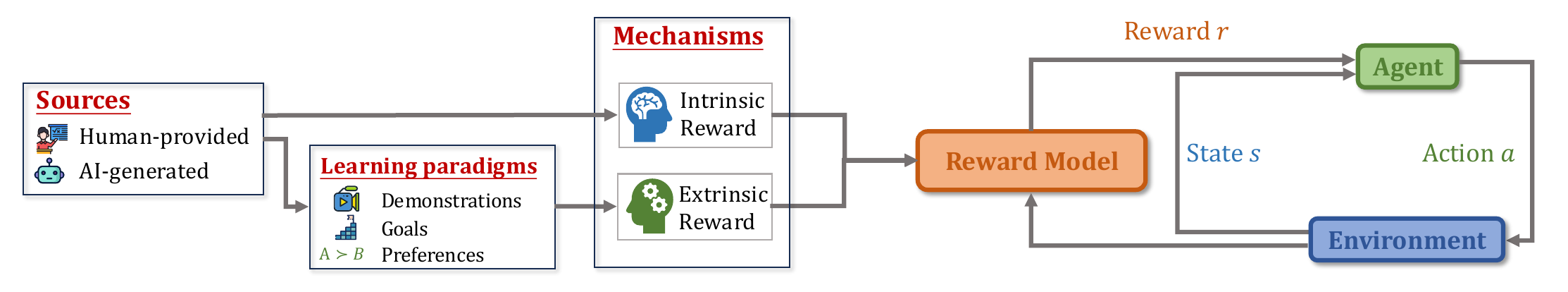}
\caption{A framework for reward modeling in RL, categorizing reward models by their sources, feedback types, and mechanisms to provide a structured understanding of how rewards are derived and utilized in RL systems.}
    \label{fig:pipeline}
\end{figure*}

\begin{enumerate}
    \item \textbf{Background of reward modeling (Section \ref{sec:background}).} We first provide the necessary background on RL and reward models;
    \item \textbf{Categorization of reward models.} We propose a classification framework for reward models, distinguishing them by three key factors: the \textit{source} (\textbf{Section \ref{sec:origin}}), the \textit{mechanism} that drives learning \textbf{(Section \ref{sec:mechanism})}, and the \textit{learning paradigm} used to derive rewards \textbf{(Section \ref{sec:method})}. We also list recent publications about reward modeling and categorize them based on our hierarchy in Table~\ref{table:category}. 
    \item \textbf{Applications and evaluation methods of reward models (Section \ref{sec:application} and Section \ref{sec:evaluation}).} We provide a discussion on the applications of reward models in practical scenarios, together with evaluation methods for these models.
    \item \textbf{Prosperous directions and discussions (Section \ref{sec:summary}).} We summarize this survey by presenting potential future directions in this topic.
\end{enumerate}

\section{Background}\label{sec:background}

RL is typically formulated as a Markov Decision Process (MDP) $\langle \mathcal{S}, \mathcal{A}, T, R, \gamma \rangle$, where $\mathcal{S}$ and $\mathcal{A}$ denote the state space and the action space, respectively. The transition function $T(\cdot|s, a)$ defines the distribution over the next states after taking action $a$ at state $s$. The reward model $R(s, a)$ specifies the instantaneous reward that the agent will receive after taking action $a$ at state $s$, and $\gamma$ is the discount factor that balances the importance of future rewards. An RL agent aims to find the policy $\pi(a|s)$ maximizing the following expected discounted cumulative reward (a.k.a. return):
\begin{equation}
    \mathcal{J}(\pi) = \mathbb{E}_{\pi,T}\left[\sum_{t=0}^{\infty}\gamma^t R(s_t, a_t)\right].
\end{equation}
where the expectation is taken over the distribution of states and actions that the agent will encounter following $\pi$ and $T$. 

The fundamental objective of learning is to refine an agent's behavior to accomplish predefined goals or tasks. While supervised learning (SL) offers a principled approach by training agents on human-annotated datasets to mimic human behavior, this method is limited by the quantity and quality of available human demonstrations. Consequently, agents trained solely by SL may make irrational decisions when human behavior is missing or sub-optimal. Reinforcement learning instead offers another principled way that permits the agent to explore the environment autonomously and adapt its behavior based on the rewards it receives. Such a trial-and-error approach exempts the agent from the constraints of datasets and opens the possibility of achieving or even surpassing human-level performance.

Although $\mathcal{S}, \mathcal{A}$, and the transition model $T$ are inherently defined by the environment, the reward model $R$ must be carefully crafted by the task designer. This careful design is crucial to ensure that the specified rewards truly reflect the underlying objectives. In many applications, only descriptive guidelines or standards of the intended goals are available, and therefore we need to convert them into statistical reward models. This process is termed as \textit{reward modeling} throughout this survey. 

\section{Sources of Rewards} \label{sec:origin}

In this section, we explore different sources of reward signals in RL. We categorize reward sources into two main types: human-provided rewards, which leverage human expertise and supervision, and AI-generated rewards, which rely on foundation models typically trained by self-supervised learning on internet-scale datasets. 

\subsection{Human-Provided Rewards}

\subsubsection{Manual Reward Engineering}

Manual reward engineering refers to the process where researchers meticulously design reward functions to steer agents toward optimal policies. Take the walker task in Gym-MuJoCo \citep{towers2024gymnasium} as an example: its reward is manually designed as a combination of survival, forward movement, and control cost penalties. However, reward engineering requires human experts to translate ambiguous task objectives into precise statistical models. Such an undertaking can be both resource-intensive and perilous: if the reward function is inadequately crafted, the agent may suffer from reward hacking, leading to unpredictable behaviors \citep{kaufmann2023survey}.

\subsubsection{Human-in-the-Loop Reward Learning}
Instead of directly crafting the reward models, human-in-the-loop reward learning derives rewards from indirect human supervision, including demonstrations \citep{abbeel2004apprenticeship}, goals \citep{liu2022goal}, and preferences \citep{kaufmann2023survey}. Compared to manual reward engineering, asking human experts to provide demonstrations or feedback of such kind is much more straightforward. However, the reward learning process needs to be specifically designed to accommodate different kinds of supervision and ensure alignment with the intended task objectives.

\subsection{AI-Generated Rewards}

Foundation models, such as large language models (LLMs) and vision-language models (VLMs) pre-trained on internet-scale human-generated data, have demonstrated a remarkable ability to interpret human intent and autonomously define reward models for RL. For instance, LLMs have been employed to design reward functions \citep{xie2023text2reward} and generate feedback for reward learning \citep{klissarov2023motif, bai2022constitutional, lee2024rlaif}. VLMs, in particular, are highly effective in specifying rewards and tasks within visually complex environments. Some studies 
\citep{fan2022minedojo, sontakke2023roboclip}
compute semantic similarity between agent states and task descriptions, enabling dense reward signals from visual observations. Others \citep{wang2024rl} utilize VLMs to analyze visual inputs and generate preference-based feedback for reward model training. While certain approaches \citep{baumli2023vision} leverage off-the-shelf foundation models for zero-shot reward specification, others \citep{fan2022minedojo, sontakke2023roboclip} fine-tune these models on domain-specific datasets to improve reward design.

\section{Reward Mechanisms}\label{sec:mechanism}
In this section, we focus on two different reward mechanisms that drive RL agent's learning.

\subsection{Extrinsic Reward}
Rewards are defined by incentives that drive the agent. The term \textit{extrinsic reward} corresponds to incentives that arise from external sources and directly relate to the desired task objective, e.g., instructions or goals set by supervisors or employers. Defining extrinsic rewards requires the task designer to translate abstract goals into concrete, quantifiable rewards that can be incorporated into a standard RL pipeline. The approach to accomplish this is detailed in Section \ref{sec:method}.

\subsection{Intrinsic Motivation}

In contrast to extrinsic rewards, intrinsic motivation (IM) captures an agent's innate motivation to explore and refine its behavior in the environment \citep{ryan2000intrinsic}. \citet{harlow1950learning} observed that even without extrinsic stimulus, monkeys have spontaneous desire and curiosity to solve complex puzzles. Later \citep{barto2004intrinsically} introduced IM into the reward mechanism, leading to the application of intrinsic reward. 
Unlike extrinsic rewards, intrinsic rewards are often disentangled from specific task objectives; rather, they encapsulate the encouragement for beneficial behaviors for problem-solving, such as exploration.

To coordinate the intrinsic reward and extrinsic reward, one common approach is to compute the agent's reward \( r \) as a weighted sum of the intrinsic reward \( r_\text{int} \) and the extrinsic reward \( r_\text{ext} \):
\begin{align}
    r = \lambda r_{\text{int}} + (1-\lambda)r_{\text{ext}},
\end{align}
where \(0\leq\lambda\leq 1\) is a coefficient that balances the intrinsic reward $r_{\text{int}}$ and extrinsic reward $r_{\text{ext}}$.

Next, we introduce three widely used types of intrinsic motivation in reinforcement learning.

\subsubsection{Exploration}

IM has long been used to encourage exploration. By leveraging concepts such as surprise \citep{pathak2017curiosity}, epistemic uncertainty \citep{houthooft2016vime}, and disagreement \citep{pathak2019self, sekar2020planning}, many methods quantify the strangeness of states as the prediction errors of state transition, and thus use the errors as intrinsic rewards to encourage the agent to explore unseen areas of the environment. The strangeness of states can also be quantified using the distillation error between randomly initialized networks \citep{burda2018exploration}, which can be more flexible to implement.

Other works design intrinsic rewards for exploration through the lens of data diversity. Among them, count-based methods, such as the well-known upper confidence bound (UCB) \citep{lai1985asymptotically}, maintain the state visitation counts and assign higher intrinsic rewards for less-visited states. Later, static hashing \citep{tang2017exploration} and density estimation \citep{bellemare2016unifying,ostrovski2017count} are incorporated to extend count-based exploration to problems with larger or even continuous state spaces. On the other hand, \citet{liu2021behavior} and \citet{badia2020never} promote diversity by estimating the data entropy and using the entropy as the intrinsic rewards. In this way, they can encourage the agent to explore novel and diverse states.

\subsubsection{Empowerment}

Empowerment, an information-theoretic intrinsic motivation (IM) concept, motivates an agent to maximize its influence on the environment by seeking states where it possesses the greatest control over future outcomes \citep{klyubin2005empowerment}. An intrinsic reward signal can then be formulated to guide the agent's exploration towards states that offer greater control and a wider diversity of achievable consequences. Many previous works leverage empowerment for skill discovery \citep{eysenbach2018diversity, mazzaglia2022choreographer}. These works aim to find a skill-conditioned policy $\pi(a | s,z)$ that maximizes the mutual information between the resulting trajectory and the latent variable z. The intrinsic reward is designed based on the decomposition of this mutual information. The agent is then encouraged to recover the latent $z$ from the trajectory, implying that different $z$ should produce distinctly different trajectories, thereby defining $z$ as the skill. By providing an intrinsic reward based on the agent's potential to influence the environment, skill learning through empowerment enables more generalizable agent behaviors and facilitates rapid adaptation to new tasks.

\subsubsection{Knowledge-Driven IM}

Many approaches leverage high-level knowledge and structured reasoning to generate intrinsic rewards, bridging the gap between abstract understanding and low-level sensorimotor interactions. Some methods derive preferences from structured event descriptions, comparing pairs of observations to infer meaningful intrinsic signals \citep{klissarov2023motif}. \citet{xu2023internal} adopted a reward-shaping technique by treating valuable propositional logic knowledge as intrinsic rewards for the RL procedure. \citet{du2023guiding} generates goal candidates based on an agent’s current context and provides rewards for achieving those inferred objectives. In recent work \citep{klissarov2023motif}, large-scale models such as LLMs and VLMs have been employed to facilitate this process due to their broad knowledge and reasoning capabilities.

\begin{table*}[t]
\centering
\begin{tabularx}{\linewidth}{cccc}
\toprule
\textbf{Source} & \textbf{Mechanism} & \textbf{Feedback} & \textbf{Method}\\
\midrule
human & intrinsic & - & \multicolumn{1}{m{0.6\linewidth}}{\citep{pathak2017curiosity, houthooft2016vime, pathak2019self, sekar2020planning, burda2018exploration, bellemare2016unifying, badia2020never, liu2021behavior, eysenbach2018diversity, mazzaglia2022choreographer, wan2024moser}}\\
\midrule
AI & intrinsic & - & \multicolumn{1}{m{0.6\linewidth}}{\citep{klissarov2023motif, xu2023internal, du2023guiding}}\\
\midrule
human & extrinsic & demonstration & \multicolumn{1}{m{0.6\linewidth}}{\citep{abbeel2004apprenticeship, ziebart2008maximum, finn2016connection, finn2016guided, fu2017learning, jeon2020regularized}}\\
\midrule
human & extrinsic & goal & \multicolumn{1}{m{0.6\linewidth}}{\citep{liu2022goal, nachum2018data, mazzaglia2024genrl, hartikainen2019dynamical, mendonca2021discovering, park2023metra, myers2024learning, wang2025founder}} \\
\midrule
AI & extrinsic & goal & \multicolumn{1}{m{0.6\linewidth}}{\citep{sontakke2023roboclip, fan2022minedojo, rocamonde2023vision}} \\
\midrule
human & extrinsic & preference & \multicolumn{1}{m{0.6\linewidth}}{\citep{christiano2017deep,kim2023preference,verma2024hindsight,knox2022models,touvron2023llama,liu2024reward,ouyang2022training,kopf2023openassistant,rafailov2023direct,song2024preference,liu2024lipo}}\\
\midrule
AI & extrinsic & preference & \multicolumn{1}{m{0.6\linewidth}}{\citep{bai2022constitutional, lee2024rlaif, wang2024rl}} \\
\bottomrule
\end{tabularx}
\caption{Summary of the algorithms mentioned in Section \ref{sec:origin}, Section \ref{sec:mechanism}, and Section \ref{sec:method}.}
\label{table:category}
\end{table*}

\section{Learning Paradigms}\label{sec:method}
In this section, we focus on the paradigms of learning the reward model $R_{\theta}$ from different kinds of human feedback. Specifically, existing literature that involves reward learning can be broadly categorized into three paradigms, namely:

\begin{itemize}
\item \textbf{Learning from demonstrations,} which extracts reward models based on demonstrations provided by human experts. This is related to inverse RL (IRL) \citep{arora2021survey}. 
\item \textbf{Learning from goals,} which derives reward models from specified goal states. This is related to goal-conditional RL (GCRL) \citep{liu2022goal}. 
\item \textbf{Learning from preferences,} which extracts reward models from human preferences among two or more trajectory segments. This is related to preference-based RL (PbRL) and reinforcement learning from human feedback (RLHF) \citep{kaufmann2023survey}. 
\end{itemize}
In each subsection, we will provide a brief overview of the established methods in each setting.

\subsection{Learning from Demonstrations}

\subsubsection{Maximum-Entropy Inverse Reinforcement Learning}
Previous approaches to IRL iteratively optimize the reward model to maximize the performance margin between demonstrations and any other policy, such that the demonstrations appear optimal under the learned reward model \citep{abbeel2004apprenticeship}. However, the IRL problem is inherently ill-posed, because multiple distinct rewards may explain the same expert behavior. A common strategy for resolving this ambiguity is to incorporate additional regularization into the learning objective. As an example, the maximum-entropy IRL (MaxEnt-IRL) framework \citep{ziebart2008maximum} introduces entropy regularization such that the expert demonstrations are drawn from the Boltzmann distribution:
\begin{equation}\label{eq:maxent-irl}
    \begin{aligned}
        p_\theta(\tau)=\frac{\exp(R_\theta(\tau))}{Z_\theta}, 
    \end{aligned}
\end{equation}
where $\tau=(s_1,a_1, \ldots, s_{|\tau|}, a_{|\tau|})$ denotes the demonstrated trajectory, and $R_\theta(\tau)=\sum_{t=1}^{|\tau|}R_\theta(s_t, a_t)$ is the cumulative reward along $\tau$. The partition function $Z_\theta$ normalizes the distribution, and it can be computed via dynamic programming in small, discrete domains \citep{ziebart2008maximum} or approximated by importance sampling in continuous settings \citep{finn2016guided}. By parameterizing the reward model $R_\theta$ as linear models or neural networks, we can perform maximum likelihood training based on observed demonstrations and obtain the reward models that explain the demonstrations. 

\subsubsection{Adversarial Reward Learning}

\citet{finn2016connection} demonstrated that the MaxEnt-IRL problem can be reformulated as a generative adversarial network (GAN) problem by employing a specifically structured discriminator. Let the generator of the trajectories and the reward model be $q_\psi(\tau)$ and $R_\theta(\tau)$ respectively, the discriminator is parameterized as:
\begin{equation}
    \begin{aligned}
        D_\theta(\tau)=\frac{\frac 1Z\exp(R_\theta(\tau))}{\frac 1Z\exp(R_\theta(\tau))+q_\psi(\tau)},
    \end{aligned}
\end{equation}
where $Z$ represents the partition function and can be estimated via importance sampling. The generator and the discriminator are trained via standard GAN losses: 
\begin{equation}\label{eq:gan_gcl}
    \begin{aligned}
        \mathcal{L}(\theta)&=\mathbb{E}_{\tau\sim \mathcal{D}_e}\left[-\log D_\theta(\tau)\right]+\mathbb{E}_{\tau\sim q}\left[-\log(1-D_\theta(\tau))\right],\\
        \mathcal{L}(\psi)&=\mathbb{E}_{\tau\sim q_\psi}\left[\log \frac{(1-D_\theta(\tau))}{D_\theta(\tau)}\right]\\
        &=\mathbb{E}_{\tau\sim q_\psi}\left[-R_\theta(\tau)\right]-\mathcal{H}(q_\psi)+\log Z,
    \end{aligned}
\end{equation}
where $\mathcal{D}_e$ denotes the expert demonstrations and $\mathcal{H}$ is the entropy. By optimizing \eqref{eq:gan_gcl}, we can effectively optimize the reward model $R_\theta$. When the optimization converges, it follows from the maximum-entropy theory that $q^*(\tau)\propto \exp (R^*(\tau))$, which exactly recovers the MaxEnt-IRL problem in \eqref{eq:maxent-irl}. However, conducting optimization over the trajectories incurs high variance, and therefore the adversarial inverse RL (AIRL) framework \citep{fu2017learning} further decomposes the problem and operates on a state-action level: 
\begin{equation}
    \begin{aligned}
        \mathcal{L}(\theta)= \mathbb{E}_{\mathcal{D}_e}\left[-\log D_\theta(s,a)\right]+ \mathbb{E}_{q_\psi}\left[-\log(1-D_\theta(s,a))\right],\\
    \end{aligned}
\end{equation}
where $D_\theta(s,a)=\frac{\exp f_\theta(s, a)}{\exp(f_\theta(s, a))+p_\psi(a|s)}$. Once training is complete, $f_\theta$ is shown to recover the optimal advantage function $A^*$, from which reward models may subsequently be extracted. 
Building on this foundation, the AIRL framework has been further extended -- for instance, to encompass a broader class of regularizations \citep{jeon2020regularized}
.

\subsection{Learning from Goals}
When our intended goals can be explicitly described or specified as a state $g\in \mathcal{S}$, the reward model can be conveniently defined based on whether the goal is achieved \citep{liu2022goal}:
\begin{align}
    R(s,g) = \mathds{1} (s \text{ accomplishes } g),
\end{align}
where $\mathds{1}$ is the indicator function. However, this binary reward structure is extremely sparse and inefficient for policy optimization, because the agent only receives a reward upon reaching the goal state, without intermediate supervision. To address this sparsity, an alternative solution is to reshape the reward as the distance between the current and the desired goal:
\begin{align}
    R(s,g) = -d(\phi(s),\psi(g)),
\end{align}
where $\phi$ and $\psi$ are mapping functions that transform the state $s$ and the goal $g$ to the same latent space, and $d(\cdot,\cdot)$ is a specific distance metric on that space. This distance-based reward provides a more nuanced measurement of the agent's progress toward the specified goal. In the below, we will introduce two commonly adopted distance metrics: spatial distance and temporal distance.

\subsubsection{Spatial Distance}
Spatial distance directly quantifies the similarity between states from the environment. Common approaches utilize measures such as the L2 distance \citep{nachum2018data}, and cosine similarity \citep{mazzaglia2024genrl} to assess the proximity between states. These metrics may be computed either in the raw state space \citep{nachum2018data}, or within a learned latent space \citep{mazzaglia2024genrl} which better captures and exploits the problem structure. 

\subsubsection{Temporal Distance} 

Other works focus on the notion of temporal distance, which conceptually assigns higher rewards to states that are \textit{temporally} closer to the goal state. For instance, approaches like \cite{hartikainen2019dynamical} and \citet{wang2025founder} train a distance metric function $d_\theta$, such that $d_\theta(s,g)$ approximates the number of time steps required for the agent to reach $g$ from $s$. Using $R=-d_\theta$ as the reward model, the agent will be guided toward states that are in the proximity of the goal. Moreover, \citep{park2023metra} frames temporal distance learning as a constrained optimization problem, maintaining a distance threshold between adjacent states while dispersing others. Recently, \citep{myers2024learning} defines a temporal distance metric based on successor features and temporal contrastive learning, which is shown to satisfy the quasi-metric property. Temporal distance offers a more grounded reward signal by effectively reflecting the agent's progress toward the goal and capturing deeper task semantics beyond visual details.

\subsubsection{Semantic Similarity}

Semantic similarity-based rewards measure how closely the agent’s current state aligns with a given goal in a shared representation space. RoboCLIP \citep{sontakke2023roboclip} computes the reward as the dot product between the text embedding of a language-specified goal and the video embedding of the agent's observed trajectory. MineCLIP \citep{fan2022minedojo} computes rewards as $R = \max\left(P_G - \frac{1}{N_T}, 0\right)$,
where \( P_G \) is the probability of the observation video matching the goal description against negatives, and \( \frac{1}{N_T} \) serves as a baseline to filter out uncertain estimates. These embeddings can be obtained from VLMs, which map multimodal inputs into a common space, allowing the agent to learn from high-level instructions or demonstrations.

\subsection{Learning from Preferences}

In many applications, obtaining human evaluations is comparatively cost-effective compared to collecting demonstrations or identifying the goal states. Consider training language models to follow instructions as an example, it is both tedious and time-consuming to require human annotators to generate template responses for every request. On the contrary, comparing agent-generated responses using metrics such as helpfulness, harmlessness, and truthfulness is considerably more straightforward. In this section, we therefore investigate methods for deriving rewards from human-annotated preferences among candidate options.

In this framework, annotators are asked to label their preferences $y$ between a pair of trajectories $(\tau^0, \tau^1)$, where $\tau=(s_1, a_1,\ldots,s_{|\tau|}, a_{|\tau|})$. A label $y=0$ means $\tau^0$ is preferred over $\tau^1$ (denoted as $\tau^0\succ \tau^1$), and $y=1$ implies the opposite. To build the connection between observed preferences and reward models, we need \textit{preference models}. A widely used example is \textit{Bradley-Terry} (BT) models~\citep{bradley1952rank}, which posit that the probability of preference can be described by a Boltzmann distribution applied to the cumulative reward: 
\begin{equation}\label{eq:bt_model}
    \begin{aligned}
        P_{\rm BT}(\tau^0\succ \tau^1;\theta)&= \frac{\exp(\sum_{(s^0_t, a^0_t)\in \tau^0}R_\theta(s^0_t, a^0_t))}{\sum_{j\in\{0,1\}}\exp(\sum_{(s^j_t, a^j_t)\in \tau^j}R_\theta(s^j_t, a^j_t))}.
    \end{aligned}
\end{equation}
To optimize the reward model $R_\theta$, we can maximize the likelihood of the observed preferences:
\begin{equation}\label{eq:pb_nll}
    \begin{aligned}
        &\mathcal{L}(\theta)=-\\
        &\sum_{(\tau^0,\tau^1,y)\in \mathcal{D}}(1-y)\log P (\tau^0\succ \tau^1;\theta)+y \log P(\tau^1 \succ \tau^0;\theta),
    \end{aligned}
\end{equation}
where $P$ is defined according to the preference model in \eqref{eq:bt_model}. After training, we can label the reward of each transition pair and subsequently employ any RL algorithm to optimize the policies \citep{christiano2017deep}. 
Alternatively, we can also directly train the policy via \eqref{eq:pb_nll} by reparameterizing the reward model through the policy in certain circumstances \citep{rafailov2023direct}. 

\subsubsection{Preference Models}

Despite its popularity in PbRL literature, BT models may not align with reality \citep{kim2023preference}. Consequently, several studies have proposed alternative preference models that more closely reflect the mechanisms underlying human preferences.  Preference Transformer \citep{kim2023preference} introduces importance weights over state-action pairs to account for the dependence on certain critical states in the trajectory: 
\begin{equation}\label{eq:pt}
    \begin{aligned}
        P_{\rm PT}(\tau^0\succ \tau^1;\theta)&= \frac{\exp(\sum_{(s^0_t, a^0_t)\in \tau^0}w_t^0R_\theta(s^0_t, a^0_t))}{\sum_{j}\exp(\sum_{(s^j_t, a^j_t)\in \tau^j}w^j_tR_\theta(s^j_t, a^j_t))},
    \end{aligned}
\end{equation}
where $j\in\{0,1\}$ and the weights $w^j_t$ are the average attention weights of the pair $(s^j_t,a^j_t)\in\tau^j$ calculated by a bi-directional attention layer. Similarly, \citet{verma2024hindsight} replaced weights in \eqref{eq:pt} with attention weights from a transformer-based transition model, thereby incorporating state importance priors from the perspective of transition models. Besides, the \textit{regret-based} models \citep{knox2022models} propose to model human preferences by the sum of optimal advantages along the trajectory, rather than the rewards:
\begin{equation}\label{eq:regretrm}
    \begin{aligned}
        P_{\rm Reg}(\tau^0\succ \tau^1)&= \frac{\exp(-\textrm{Regret}(\tau^0))}{\exp(-\textrm{Regret}(\tau^0))+\exp(-\textrm{Regret}(\tau^1))},\\
        \textrm{Regret}(\tau)&=\sum_{t=1}^{|\tau|}[Q^*_R(s_t,a_t)-V^*_R(s_t)],\\
    \end{aligned}
\end{equation}
with $V^*_R$ and $Q^*_R$ being the optimal state value function and Q-value function for the reward model $R$, respectively. \citet{knox2022models} demonstrated that this approach may better predict real human preference and the learned reward model may achieve superior performance in practice. 

\subsubsection{Extension to Ordinal Feedback}

Ordinal feedback generalizes binary feedback by requiring annotators to additionally specify the strengths of their preferences (e.g., \textit{slightly better} or \textit{significantly better}). To integrate this more nuanced information, existing studies modify BT models by incorporating soft margins \citep{touvron2023llama} or soft labels $y_i\in[0, 1]$ \citep{liu2024reward}, where the margin or the label reflects the strength of the preference. 

\subsubsection{Beyond Pairwise Comparisons}

Human feedback can also be provided in the form of rankings among multiple candidates \citep{ouyang2022training,kopf2023openassistant}. Although such listwise comparisons put a greater burden on annotators, they also carry richer information than pairwise comparisons. To accommodate rankings, \textit{Plackett-Luce} (PL) models \citep{plackett1975analysis} generalize BT models by extending the comparison to $K$ candidates:
\begin{equation}\label{eq:pl_model}
    \begin{aligned}
        &P_{\rm PL}(\tau^1\succ\tau^2\succ\ldots\succ \tau^K)\\
        &=\prod_{k=1}^K\frac{\exp(\sum_{(s^k_t,a^k_t)\in\tau^k}R(s^k_t,a^k_t))}{\sum_{j=k}^K\exp(\sum_{(s^j_t,a^j_t)\in\tau^j}R(s^j_t,a^j_t))},
    \end{aligned}
\end{equation}
where $(\tau^1\succ\tau^2\succ\ldots\succ \tau^K)$ is the observed ranking. Substituting \eqref{eq:pl_model} into \eqref{eq:pb_nll} yields the objective of learning rewards from rankings \citep{rafailov2023direct,song2024preference}. Another straightforward approach to rankings is breaking the ranking into pairs by selecting two candidates from the list and assigning the label according to their ranks, thereby reducing the problem of applying BT models to all possible pairwise comparisons \citep{ouyang2022training,liu2024lipo}. 

\section{Applications} \label{sec:application}

Reward model designing constitutes an indispensable step before any practical applications of RL. Therefore, in this section, we briefly review successful applications of reward models in deep RL, including control problems, generative model finetuning, and other fields. 

\subsection{Control Problems}

Reward models play a pivotal role in control problems, as a fundamental mechanism for guiding decision-making in dynamic environments. \cite{christiano2017deep} demonstrated their effectiveness in facilitating policy learning across diverse domains, including game-playing and simulated continuous control tasks. In gameplay scenarios, \cite{fan2022minedojo} leveraged generated rewards to enhance learning in Minecraft tasks. In robotics, \cite{sontakke2023roboclip} employed reward models to train agents across various robotic tasks. Similarly, in autonomous driving, the design of reward functions remains a critical aspect of training intelligent agents \citep{knox2023reward}.

\subsection{Generative Model Post-training}

Modern generative models typically feature a two-stage training procedure, where the \textit{pre-training} stage involves unsupervised learning on internet-scale data, and the \textit{post-training} stage fine-tunes the models and fits them for downstream tasks. A prominent example is InstructGPT \citep{ouyang2022training}, which employs RL to optimize model outputs based on human preference data. Specifically, it trains a reward model on human-ranked responses and fine-tunes the language model to maximize this reward. This approach has become a standard method for enhancing the helpfulness, harmlessness \citep{dai2023safe}, and general task-solving capabilities of LLMs \citep{abramson2022improving}. In mathematical problem-solving, golden rewards can be defined by comparing the model-generated answers with ground-truth answers \citep{luong2024reft} or by verifying the correctness using formal solvers \citep{xin2024deepseek}. Some works also use LLM-based verifiers \citep{zhang2024generative}, further leveraging the in-context learning ability provided by LLMs. 

\subsection{Other Fields}

In recommendation systems, \citet{xue2023prefrec} trained reward models to allow RL recommendation systems to learn from users' historical behaviors. \citet{kim2020automatic} used a reward model to automate peer-to-peer (P2P) energy trading, while \citet{oueida2019smart} designed a reward model to improve the management of healthcare resources.

\section{Evaluating Reward Models}\label{sec:evaluation}

Once reward models are developed, reliable evaluation techniques are essential for comparing or selecting models for downstream policy optimization. However, due to the ambiguous link between reward models and final policy performance, relying on a single evaluation perspective is often insufficient \citep{arora2021survey}. We categorize commonly used reward evaluation techniques into the following three types, which are often used in combination to achieve a more comprehensive assessment of reward models.

\subsection{Evaluation via Policy Performance} 

Reward model quality can be evaluated by measuring the performance of policies trained with it. Primary metrics include ground-truth reward, task success rate, and training efficiency, with superior reward models yielding higher values across these measures.
This approach is widely adopted in reinforcement learning literature to assess the alignment between reward models and actual objectives \citep{christiano2017deep}. However, these metrics are sensitive to policy optimization algorithms and environmental stochasticity, potentially limiting their ability to independently reflect the true performance of the reward model itself.

\subsection{Evaluation via Distance Metrics}

To evaluate and compare reward models, another approach is to design distance metrics that accurately reflect the behavioral differences between the policies induced by these rewards. The pioneering work EPIC \citep{gleave2020quantifying} introduces \textit{canonically shaped rewards} to remove ambiguity and invariances from reward models and proposes to use the Pearson coefficient between two canonically shaped rewards as a measure of  the reward similarity. The EPIC distance between two reward models is demonstrated to upper-bound the performance difference between the induced policies. 
Lower EPIC distances to the ground truth reward indicate superior reward modeling capability.

Based on EPIC, \citet{wulfe2022dynamics} further incorporates the dynamics information when considering the invariant reward shaping and introduces the DARD metric, which is more predictive and accurate in quantifying the differences in rewards. Furthermore, \citet{skalse2023starc} presented a general framework for designing such distance metrics. The STARC metrics provided in this framework are shown to induce both the upper bound and the lower bound of the performance differences, and any other metrics that possess the same property must be equivalent to the STARC metrics up to bilipschitz scaling. When datasets containing ground-truth rewards are available, distance metrics are particularly suitable for offline evaluation, circumventing the necessity for policy learning.

\subsection{Evaluation via Interpretable Representations}

Although the evaluation of reward models may not be straightforward, we can transform them equivalently into interpretable representations. \citet{jenner2022preprocessing} proposed to transform reward models with potential-based shaping and visualize the shaped reward instead. Since potential-based shaping preserves the optimal policy, characteristics of the shaped reward may also apply to the original reward model. Alternatively, we can evaluate the reward model through the behavior of the induced policy \citep{rocamonde2023vision}.

\section{Conclusions} \label{sec:summary}

Recently, reward models have become a highly motivating area of research, driven by both theoretical challenges and practical needs across various domains. We consider the development of reward models as a significant step before the application of RL to real-world problems, and we hope this survey can offer valuable insights for both researchers and practitioners. Although our study provides a comprehensive overview of the topic, the design and variations of reward models still extend beyond the scope of this discussion. Interested readers can also refer to other survey papers 
\citep{eschmann2021reward, liu2022goal, arora2021survey, kaufmann2023survey} that focus on RL subfields closely related to reward modeling.

\subsection{Future Directions}
Efficient and accurate reward modeling is a valuable research direction with significant application prospects. It combines increasingly mature technologies such as large models and diffusion models with reward design and generation in reinforcement learning to provide behavioral feedback for agents in perception, planning, decision-making, and navigation. Although there is no definitive conclusion on which route can achieve efficient reward modeling, research on various technologies in recent years has effectively promoted the development of this field. With the continuous development of machine learning and reinforcement learning, reward modeling has many valuable research directions in the future, including:

\begin{enumerate}
    \item \textbf{Vectorized rewards}: Constructing vectorized rewards to replace scalarized single rewards, dynamically balancing multiple competitive reward signals to provide agents with more comprehensive feedback. 
    \item \textbf{Interpretating reward models}: Improving the transparency of reward functions and explaining the decision-making logic behind reward models. 
    \item \textbf{Ethical alignment and social value constraints}: Quantifying ethical principles and embedding them into reward functions while avoiding potential side effects during the optimization process. 
    \item \textbf{Reward foundation models}: Similar to constructing a general representation space, consider training a foundation reward model that can obtain general reward values based on diverse inputs (such as limb movements).
\end{enumerate}

\section*{Acknowledgements}

We thank the anonymous reviewers for their insightful feedback. This work was supported by the National Science and Technology Major Project (Grant No. 2022ZD0114805) and Young Scientists Fund of the National Natural Science Foundation of China (PhD Candidate) (Grant No. 624B200197).

\small
\bibliographystyle{named}
\bibliography{ijcai25}

\end{document}